\documentclass[conference]{IEEEtran}
\IEEEoverridecommandlockouts
\def\BibTeX{{\rm B\kern-.05em{\sc i\kern-.025em b}\kern-.08em
    T\kern-.1667em\lower.7ex\hbox{E}\kern-.125emX}}

\usepackage[utf8]{inputenc} % allow utf-8 input
\usepackage[T1]{fontenc}    % use 8-bit T1 fonts
\usepackage{hyperref}       % hyperlinks
\usepackage{url}            % simple URL typesetting
\usepackage{booktabs}       % professional-quality tables
\usepackage{amsfonts}       % blackboard math symbols
\usepackage{nicefrac}       % compact symbols for 1/2, etc.
\usepackage{microtype}      % microtypography
\usepackage{xcolor}         % colors

\usepackage[utf8]{inputenc} % allow utf-8 input
\usepackage[T1]{fontenc}    % use 8-bit T1 fonts
\usepackage{hyperref}       % hyperlinks
\usepackage{url}            % simple URL typesetting
\usepackage{booktabs}       % professional-quality tables
\usepackage{amsfonts}       % blackboard math symbols
\usepackage{nicefrac}       % compact symbols for 1/2, etc.
\usepackage{microtype}      % microtypography

\usepackage{amssymb, amsthm, amsmath, amsfonts, mathtools}
\usepackage{xfrac}
\usepackage{graphicx}
\usepackage{subfigure}
\usepackage{wrapfig}
\usepackage{color}
\usepackage{listings}
\usepackage{makecell}
\usepackage{multirow}
\usepackage[usestackEOL]{stackengine}
\usepackage{setspace}
\usepackage{listings}
\usepackage{algorithm}
\usepackage{algorithmic}

\usepackage{tikz}
\usepackage{pgfplots}
\usepackage{pgfplotstable}
\usetikzlibrary{positioning,shapes, fit}
\pgfplotsset{table/search path={data}}
\pgfplotsset{compat=1.15}

\pgfplotscreateplotcyclelist{thetalist}{
	{cyan!60!black,mark=none},
	{red,mark=none},
	{blue,mark=none},
	{orange,mark=none},
	{black,mark=none}
}

\definecolor{pyplot1}{RGB}{31, 119, 180}
\definecolor{pyplot2}{RGB}{255, 127, 14}
\definecolor{pyplot3}{RGB}{44, 160, 44}
\definecolor{pyplot4}{RGB}{214, 39, 40}
\definecolor{pyplot5}{RGB}{148, 103, 189}
\definecolor{pyplot6}{RGB}{140, 86, 75}
\definecolor{pyplot7}{RGB}{227, 119, 194}
\definecolor{pyplot8}{RGB}{127, 127, 127}
\definecolor{pyplot9}{RGB}{188, 189, 34}
\definecolor{pyplot10}{RGB}{23, 190, 207}
\pgfplotscreateplotcyclelist{pyplot}{
	{pyplot1},
	{pyplot2},
	{pyplot3},
	{pyplot4},
	{pyplot5},
	{pyplot6},
	{pyplot7},
	{pyplot8},
	{pyplot9},
	{pyplot10}
}

\pgfplotsset{
compat=1.11,
legend image code/.code={
\draw[mark repeat=2,mark phase=2] plot coordinates { (0cm,0cm) (0.15cm,0cm) (0.3cm,0cm) };
}}
\usepackage{subfiles} % Best loaded last in the preamble

\newcommand{\E}[2][]{\mathbb{E}_{{#1}} \left[ {#2} \right] }

\DeclareMathOperator{\Tr}{Tr}

\newtheorem{definition}{Definition}
\newtheorem{theorem}{Theorem}    
    
\begin{document}

\title{Dissipative residual layers for unsupervised implicit parameterization of data manifolds\\
%\title{Conference Paper Title*\\
%\thanks{}
%\thanks{This manuscript has been authored by UT-Battelle, LLC under Contract No. DE-AC05-00OR22725 with the U.S. Department of Energy.  The publisher, by accepting the article for publication, acknowledges that the U.S. Government retains a non-exclusive, paid up, irrevocable, world-wide license to publish or reproduce the published form of the manuscript, or allow others to do so, for U.S. Government purposes. The DOE will provide public access to these results in accordance with the DOE Public Access Plan (http://energy.gov/downloads/doe-public-access-plan)}
}

%\author{\IEEEauthorblockN{1\textsuperscript{st} Author}
%\IEEEauthorblockA{\textit{} \\
%\textit{}\\
% \\
%\\
%}

\author{\IEEEauthorblockN{1\textsuperscript{st} Viktor Reshniak}
\IEEEauthorblockA{\textit{Data Analysis and Machine Learning Group} \\
\textit{Oak Ridge National Laboratory}\\
Oak Ridge, TN 37831 \\
reshniakv@ornl.gov\\
https://orcid.org/0000-0003-1545-4462}
%\and
%\IEEEauthorblockN{1\textsuperscript{nd} Given Name Surname}
%\and
%\and
%\and
%\and
}

\maketitle

\begin{abstract}
We propose an unsupervised technique for implicit parameterization of data manifolds.
In our approach, the data is assumed belonging to a lower dimensional manifold in a higher dimensional space, and the data points are viewed as the endpoints of the trajectories originating outside the manifold.
Under this assumption, the data manifold is an attractive manifold of a dynamical system to be estimated.
We parameterize such dynamical system with a residual neural network and propose a spectral localization technique to ensure it is locally attractive in the vicinity of data.
We also present initialization and additional regularization of the proposed residual layers. % that we call dissipative bottlenecks. 
We mention the importance of the considered problem for the tasks of reinforcement learning and support our discussion with examples demonstrating the performance of the proposed layers in denoising and generative tasks.
\end{abstract}


\section{Introduction}

Overparameterization has proven to be a blessing rather than a curse of modern neural network architectures.
It is essential for finding nearly-global optimal solutions by reducing the number of spurious local minima \cite{Safran2018,AllenZhu2019,Oymak2019}, and has been proven to be necessary for learning models that are both robust and accurate \cite{bubeck2020a,bubeck2021a}.
Empirical observations support these result by demonstrating that high capacity is required to defend against strong adversaries in practice \cite{madry2018towards}.
%This has also been last result is supported empirically by results demonstrating that high capacity is required to defend against strong adversaries \cite{madry2018towards}.

Of course, overparameterization itself is not sufficient to guarantee stable and predictable behavior of trained models.
Careless application of deep networks to out-of-distribution and unforeseen data can easily lead to catastrophic behavior hindering their use in safety critical tasks.
For example, in reinforcement learning, the ability of models to safely recover from exploratory behavior and random perturbations is required to learn stable optimal policies \cite{khalil2015nonlinear, berkenkamp2017safe}.
It would be beneficial if one has a non-intrusive approach that can be easily added to an existing architecture and improve the stability of a model without altering its behavior on the observed experimental data.
Here we make a small step towards this goal and propose an end-to-end-trainable layer for the implicit parameterization of data as a stable manifold of a dynamical system.
%In our approach, we employ the dynamical systems reformulation of deep neural networks (DNNs) to interpret the evolution of hidden states as a flow in phase space.
%The benefits of overparameterization remain valid for dynamical systems reformulations of DNNs. % as well. % such as residual networks and Neural ODEs.
%% by connecting the nu.
%%In this case, one can connect the number of parameters to the
%Moreover, this connection brings another perspective to interpret the evolution of hidden states as a flow in phase space.
The parameter estimation in this case is given by the optimal control problem~\cite{E2019}
\begin{align}\label{eq:optimal_control}
    &\min_{\gamma_t} \E[\mu]{L\big(y_T,f(x)\big) + \int_0^T R\big(\gamma_t,y_t\big)dt},
    \\\label{eq:state_equation}
    &\text{subject to }\quad y_t = F\big( \gamma_t, x \big),
\end{align}
where $L\big(y_T,f(x)\big)$ is a terminal loss function, $R\big(\gamma_t,y_t\big)$ is a regularizer, $\mu$ is a probability distribution of the input-target data pairs $(x,f(x))$, and the evolution of $y_t$ in phase space is driven by a flow $F:\mathbb{R}^p\times\mathbb{R}^d\to\mathbb{R}^d$ through the input data $y_0=x\in\mathbb{R}^d$.
Typical examples include residual layers \cite{He2016} or neural ODEs \cite{Chen2018}, i.e.,
\begin{align*}
    %\label{eq:residual_layer}
    y_t = y_{t-1} + F\big( \gamma_t, y_{t-1} \big)
    \qquad\text{and}\qquad
    \dot{y}_t = F\big( \gamma_t, y_{t-1} \big),
\end{align*}
and their various extensions. % such as stable, implicit and reversible architectures \cite{Haber_2017, Gomez2017, reshniak2020robust} \textbf{[More references]}.

Overparameterization benefits dynamical system reformulation of neural networks as well.
Dupont et al. \cite{Dupont2019augm} showed that Neural ODEs preserve the topology of the input space implying that certain functions cannot be represented by continuous flows.
%Given the form of parameterization $\gamma_t\in\mathbb{R}^p$, the size of the network in \eqref{eq:state_equation} is determined by the dimension $d$ of the phase space and the final time horizon $T$.
%Some initial results on the impact of $T$ on the approximation power of DNNs has been studied, for example, in \cite{esteve2020large,Faulwasser2021a} using classical turnpike properties of optimal control problems. % based on reachability assumptions and dissapitivity-inducing choice of regularization terms.
%It was shown that, under certain assumptions, the training error is on the order of $O(T^{-1})$.
%The impact of $d$ has been studied in \cite{Dupont2019augm}, and it was shown that Neural ODEs preserve the topology of the input space implying that certain functions cannot be represented by continuous flows.
Augmenting the original space with additional dimensions allowed to resolve this issue and lead to models that generalized better and produced simpler flows with a lower computational cost. % even when such augmentation was not required by the topological argument.
%This observation is in line with the claim that networks with excessive degrees of freedom can generalize better and are easier to train.
In fact, many DNNs can be viewed as dynamical systems embedded in a higher dimensional space.
For example, the number of pixels in images commonly exceeds the dimension of the corresponding image manifolds facilitating efficient out-of-manifold adversarial attacks \cite{lin2020dual}.

Robust/adversarial training with data augmentation reduces standard accuracy and generalizes poorly to unforseen data perturbations \cite{tsipras2018robustness, Song2018adv}.
In this effort, we propose an unsupervised technique for implicit parameterization of data manifolds that does not require explicit data augmentation.
In our approach, the data is assumed belonging to a lower dimensional manifold in a higher dimensional space, and the data points are viewed as the endpoints of the trajectories originating outside the manifold.
Under this assumption, the data manifold is an attractive manifold of a dynamical system to be estimated.
For the representation of such dynamical system, we introduce dissipative residual layers with explicit Lipschitz constants and a spectral localization approach that allows for the precise control of the spectrum of such layers.
The proposed layer is end-to-end trainable and can be easily added into existing network architectures.

%Our original motivation for this work was to propose a method to stabilize deep latent dynamics that can improve the robustness of deep models without data augmentation. 
%The current effort is only a step towards that goal.
%Nevertheless, we found the obtained results interesting enough to be shared separately.

\section{Dissipative residual layers}

\subsection{Definitions}

%\begin{align}\label{eq:our_layer}
%    y = x + \Phi(\gamma,x,y).
%\end{align}

Following the definition proposed in \cite{reshniak2020robust}, consider the trajectory generated by the implicit residual layer
%Following the definition of implicit residual layers proposed in \cite{reshniak2020robust}, consider its extension given by the autonomous block of such layers
%Consider the following autonomous block of implicit layers extending~\eqref{eq:our_layer}
%Additionally, instead of a single layer in \eqref{eq:our_layer}, one might consider a block of implicit layers on a given ``time interval" $t\in[0,T]$
\begin{align}\label{eq:layer_block}
%    &y_0 = x,
%    \\ \nonumber
%    &y_t = y_{t-1} + F(\gamma,(1-\theta)y_{t-1}+\theta y_t), \qquad t=1,\hdots,T.
	%RL[F,\theta,T](x):
	%y_T=RL_{T,\theta}(x):
%	\Phi_{\theta}(x):
%	\qquad
	y = x + F(\gamma,(1-\theta)x+\theta y).
%    \begin{cases}
%	    y_0 = x,
%	    \\
%	    y_t = y_{t-1} + F(\gamma,(1-\theta)y_{t-1}+\theta y_t), \qquad t=1,\hdots,T.
%    \end{cases}
\end{align}
%Additionally, consider 
and another perturbed trajectory given by $$\overline{y}=\overline{x}+F(\gamma,(1-\theta)\overline{x}+\theta \overline{y}).$$ 
By linearizing $F(\gamma,\cdot)$ around the reference trajectory, the evolution of the perturbation $\overline{y}-y$ is described by the linear system
\begin{align*}
	\overline{y}-y
	&= \left(I-\theta \frac{\partial F(\gamma,{z})}{\partial {z}} \right)^{-1} \left( I+(1-\theta)\frac{\partial F(\gamma,{z})}{\partial {z}} \right) (\overline{x}-x)
	\\	
	&= \mathcal{R}_{\theta}\left(\frac{\partial F(\gamma,{z})}{\partial {z}}\right) (\overline{x}-x),
	\qquad
	{z} = (1-\theta){x}+\theta {y}.
\end{align*}

The eigenvalues of the Jacobain $D_{{z}}F(\gamma,{z})$ determine the local stability properties of the system in~\eqref{eq:layer_block}.
This can be seen by looking at the Jordan normal form $J$ of $D_{{z}}F(\gamma,{z})=PJP^{-1}$ such that
\begin{align*}
	\mathcal{R}_{\theta}\left(\frac{\partial F(\gamma,{z})}{\partial \overline{z}}\right)
	= \mathcal{R}_{\theta}\left(PJP^{-1}\right)
	= P\mathcal{R}_{\theta}(J)P^{-1}.
\end{align*}
The diagonal of the matrix $J$ contains the eigenvalues of the Jacobian $D_{{z}}F(\gamma,{z})$, and the diagonal of the matrix $\mathcal{R}_{\theta}(J)$ is obatined by evaluating the matrix function $\mathcal{R}_{\theta}$ elementwise on the diagonal entries of $J$.
In the numerical analysis of ordinary differential equations, e.g. \cite{Hairer1993}, the corresponding scalar function $\mathcal{R}_{\theta}$ is known as the \textbf{stability function} of \eqref{eq:layer_block}. It is hence given by 
\begin{align}\label{eq:stab_function}
	\mathcal{R}_{\theta}(z) = \frac{1+(1-\theta)z}{1-\theta z}, \quad z\in\mathbb{C}
\end{align}
%
%It is known from the numerical analysis of ordinary differential equations, e.g. \cite{Hairer1993}, that the \textbf{stability function} of \eqref{eq:layer_block} is given by 
%\begin{align}\label{eq:stab_function}
%	\mathcal{R}_{\theta}(z) = \frac{1+(1-\theta)z}{1-\theta z}, \quad z\in\mathbb{C}.
%\end{align}
and describes the evolution of the eigenmodes of the linear perturbation to the solution of \eqref{eq:layer_block}, i.e.,
\begin{align*}
	\overline{x}-x \in span(v_1,...,v_d)
	%\quad\to\quad
	%%\overline{y}-y \in span\Big(\mathcal{R}_{\theta}(\lambda_1) v_1,...,\mathcal{R}_{\theta}(\lambda_d) v_d\Big) ,
\end{align*}
implies
\begin{align*}
	\overline{y}-y \in span\Big(\mathcal{R}_{\theta}(\lambda_1) v_1,...,\mathcal{R}_{\theta}(\lambda_d) v_d\Big) ,
\end{align*}
where $\lambda_i$, $v_i$ are the eigenvalues and (generalized) eigenvectors of the Jacobian $D_{{z}}F(\gamma,{z})$. %|_{y=x}$.

\begin{definition}\label{def:stab_region}
\textbf{Stability region} is a set of all $z\in\mathbb{C}$ such that $|\mathcal{R}_{\theta}(z)|<1$.
It contains those eigenvalues of the Jacobian $D_zF(\gamma,z)$ corresponding to stable subspaces of the linearized system.
Perturbations along such stable directions asymptotically vanish.
\end{definition}

\begin{definition}\label{def:dissip_layer}
If all eigenvalues of the Jacobian evaluated at any arbitrary point are within the stability region of a layer, than it is unconditionally stable an we call it a \textbf{dissipative residual layer}.
\end{definition}

\begin{figure}[t]
    \centering
	\stackinset{c}{0pt}{t}{-12pt}{\small $\theta=0.00$}{ \includegraphics[width=0.4\linewidth]{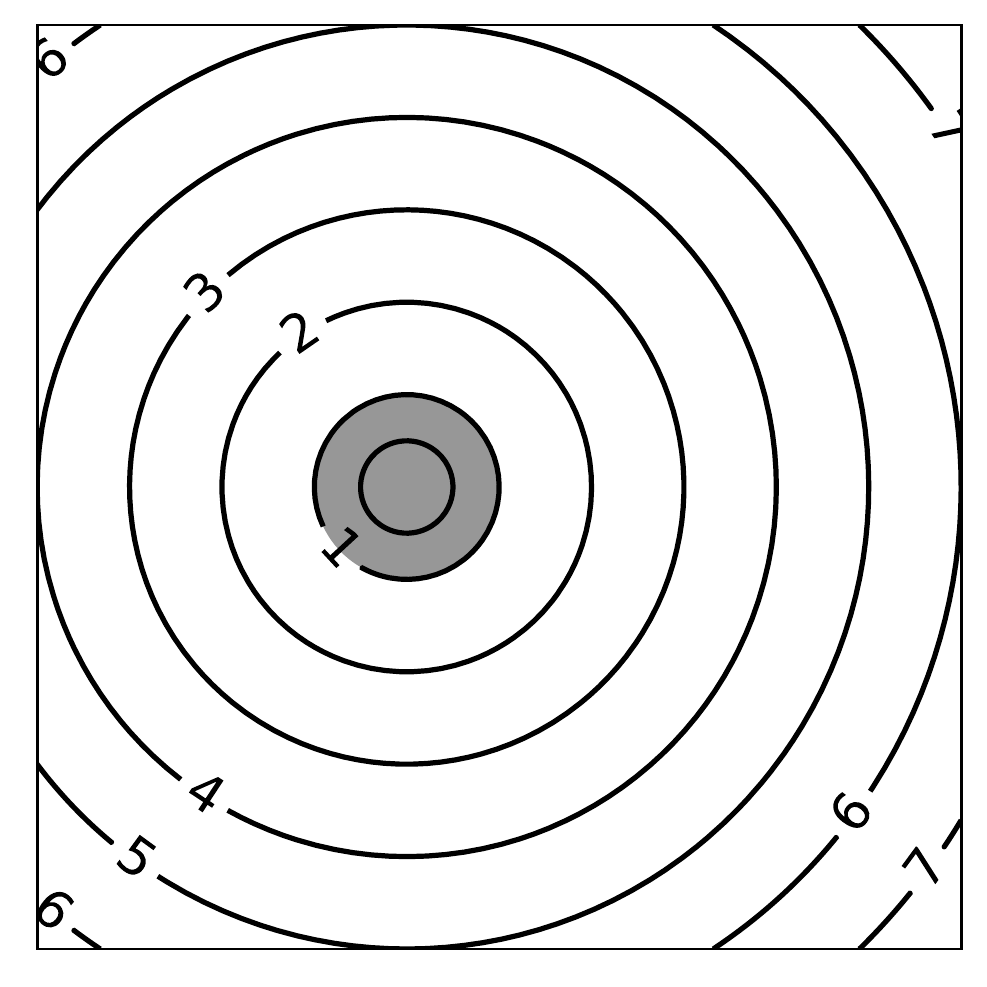}}
	\stackinset{c}{0pt}{t}{-12pt}{\small $\theta=0.25$}{ \includegraphics[width=0.4\linewidth]{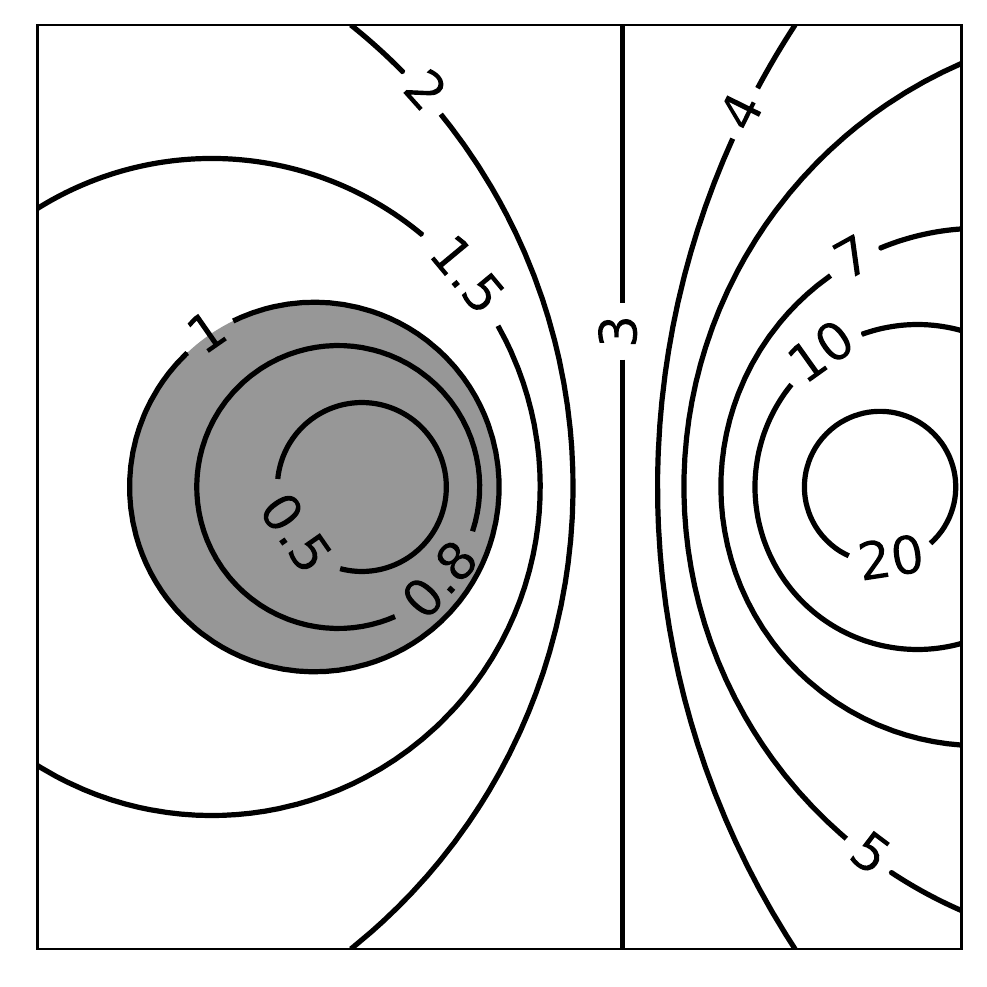}}
	\\[0.5em]
	\stackinset{c}{0pt}{t}{-12pt}{\small $\theta=0.50$}{ \includegraphics[width=0.4\linewidth]{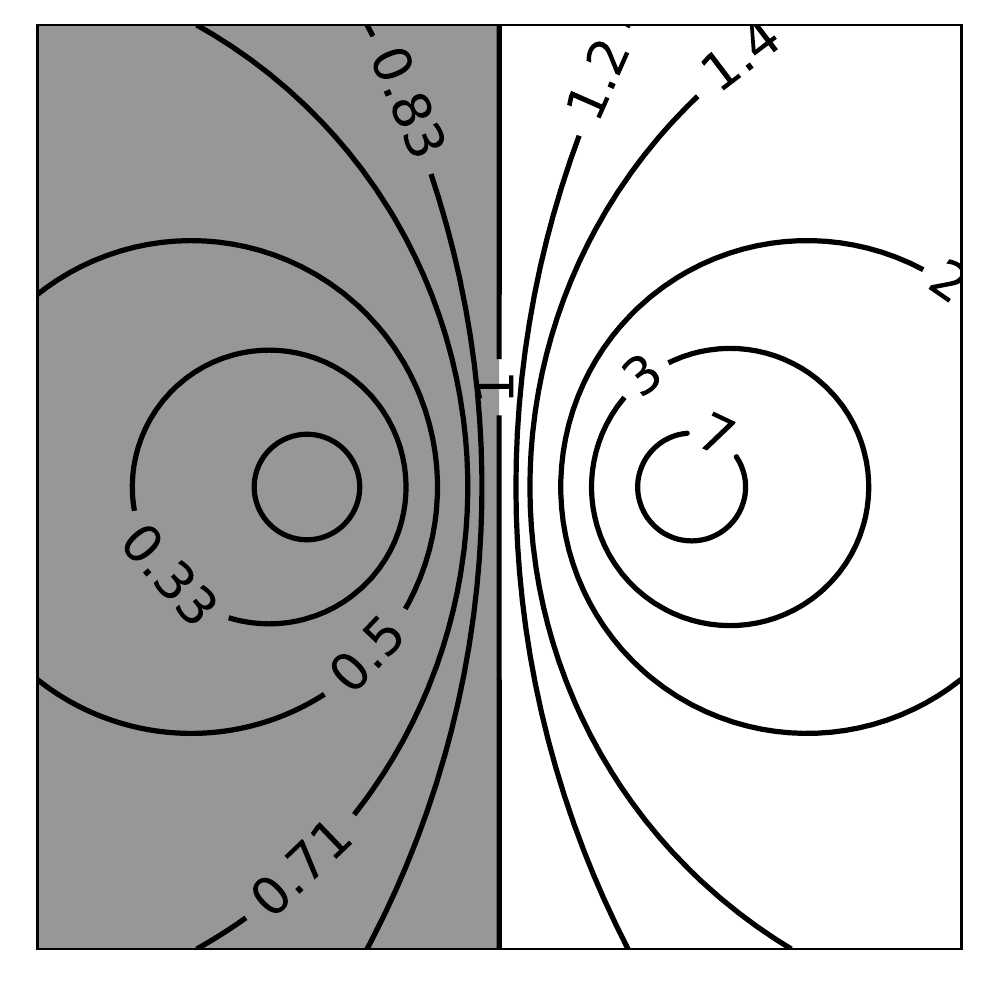}}
%	\stackinset{c}{0pt}{t}{-12pt}{\small $\theta=0.75$}{ \includegraphics[width=0.4\linewidth]{images/stab_075.pdf}}
	\stackinset{c}{0pt}{t}{-12pt}{\small $\theta=1.00$}{ \includegraphics[width=0.4\linewidth]{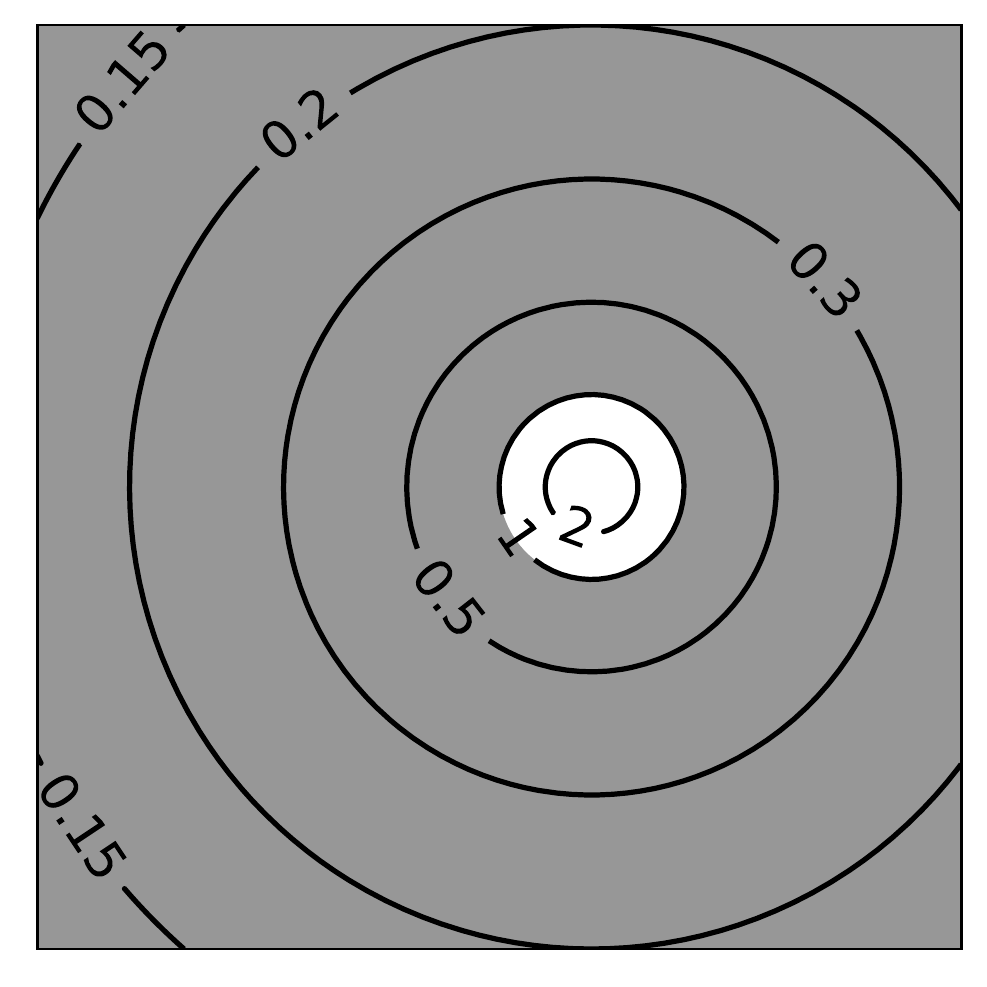}}
    \caption{Stability regions (grey) and contours of $|\mathcal{R}_{\theta}(z)|$.} % in \eqref{eq:stab_function}.}
    \label{fig:stab_regions}
\end{figure}

Figure~\ref{fig:stab_regions} illustrates stability regions of several implicit residual layers.

%\begin{figure*}[t]
%    \centering
%	\stackinset{c}{0pt}{t}{-12pt}{\small $\theta=0.00$}{ \includegraphics[width=0.18\linewidth]{images/stab_000.pdf}}
%	\stackinset{c}{0pt}{t}{-12pt}{\small $\theta=0.25$}{ \includegraphics[width=0.18\linewidth]{images/stab_025.pdf}}
%	\stackinset{c}{0pt}{t}{-12pt}{\small $\theta=0.50$}{ \includegraphics[width=0.18\linewidth]{images/stab_050.pdf}}
%	\stackinset{c}{0pt}{t}{-12pt}{\small $\theta=0.75$}{ \includegraphics[width=0.18\linewidth]{images/stab_075.pdf}}
%	\stackinset{c}{0pt}{t}{-12pt}{\small $\theta=1.00$}{ \includegraphics[width=0.18\linewidth]{images/stab_100.pdf}}
%    \caption{Stability regions (grey) and contours of $|\mathcal{R}_{\theta}(z)|$.} % in \eqref{eq:stab_function}.}
%    \label{fig:stab_regions}
%\end{figure*}

\subsection{Spectral normalization and localization}

We assume that the vector field $F(\gamma,x)$ is a composition of affine maps and element-wise activations, i.e,
\begin{align*}
    F(\gamma,x) = \phi_n\circ\phi_{n-1}\circ...\circ\phi_1\circ x
    %\quad\text{with}\quad
    %\phi_i\circ x = \sigma(\gamma_i\circ x + b_i).% \quad\text{s.t.}\quad \|\sigma\|\leq 1.
\end{align*}
with
\begin{align*}
    \phi_i\circ x = \sigma(\gamma_i\circ x + b_i).% \quad\text{s.t.}\quad \|\sigma\|\leq 1.
\end{align*}

The Lipschitz constant of $F(\gamma,x)$ is given by
\begin{align*}
    Lip(F) := \sup_{x} \left\| \frac{\partial F(\gamma,x)}{\partial x} \right\|
    \leq |\sigma|^n \prod_{i=1}^n \| \gamma_i \|,
    %\qquad |\sigma|:=\sup_{x}|\sigma'(x)|,
%    \qquad\qquad
%    \tilde{\gamma}_i = \frac{\gamma_i}{\|\gamma_i\|}
%    \;\rightarrow\;
%    Lip(F)\leq 1
\end{align*}
where $|\sigma|:=\sup_{x}|\sigma'(x)|$ and $\|\cdot\|$ is the spectral norm.
Hence by setting $$\tilde{\gamma}_i = \frac{\gamma_i}{|\sigma|\cdot\|\gamma_i\|},$$ we get $Lip(F)\leq 1$ \textbf{uniformly in~$x$}.
Note that for many standard activation functions, including ReLU, hyperbolic tangent, and sigmoid, $|\sigma|=1$.

%The key component of the normalization procedure is the fast computation of the spectral norm of weight tensors $\gamma_i$.
%In practice, spectral norm can be efficiently computed for any linear operator, including convolutional operators, using power iterations. 
%%\textbf{Elaborate on this} % as was done, for instance, in \cite{reshniak2020robust}.
%It is worth noting, however, that default implementations of the spectral normalization in standard deep learning frameworks might produce incorrect results for convolutional layers when normalization is performed on the kernel rather than the linear operator itself.
%%This will almost inevitably lead to the failure of the proposed approach.

\begin{figure}[t]
 	\centering
	\pgfplotsset{every tick label/.append style={font=\scriptsize}}
	\pgfplotsset{every axis label/.append style={font=\footnotesize}}
	\pgfplotsset{every axis title/.append style={font=\footnotesize}} 	
 	
 	\foreach \n/\c/\L in {lip/1/3} %, dissip/0/1}
 	{
		\foreach \th/\tht in {000/0.00, 100/1.00}
		{
			\begin{tikzpicture}[trim axis left,trim axis right]
 				\begin{axis} [
 					width=0.36\textwidth,
 					ymode = normal,
 					xticklabels={,,},
 					yticklabels={,,},
 					enlargelimits=false,
 					axis equal image,
 					axis on top,
 					axis line style={draw=none},
 					title=\ifthenelse{\equal{\n}{lip}}{$\theta=\tht$}{},
 					ylabel=\ifthenelse{\equal{\th}{000}}{$\hat{c}=\c$, $L=\L$}{},
 					tick style={draw=none},
 					very thick ]
 					\addplot graphics [points={(-5,-5) (5,5)}, includegraphics={trim=0 0 0 0,clip}] {{{images/\n_stab_\th}.pdf}};
 				\end{axis}
 			\end{tikzpicture}
 		}
 		\\%[-2em]
	}
	\caption{Contours of $|\mathcal{R}_{\theta}|$ with disks $\mathcal{B}\Big(\mathcal{R}^{-1}_{\theta}(\hat{c}),\mathcal{R}^{-1}_{\theta}(L)-\mathcal{R}^{-1}_{\theta}(\hat{c})\Big)$; yellow and red points correspond to $\mathcal{R}^{-1}_{\theta}(\hat{c})$ and $\mathcal{R}^{-1}_{\theta}(L)$ respectively. Note that $\sup_{\lambda\in\mathcal{B}}|\mathcal{R}_{\theta}(\lambda)|=|\mathcal{R}_{\theta}({\color{red}\bullet})|=L$.}
    \label{fig:stab_regions_with_circle} 
\end{figure}
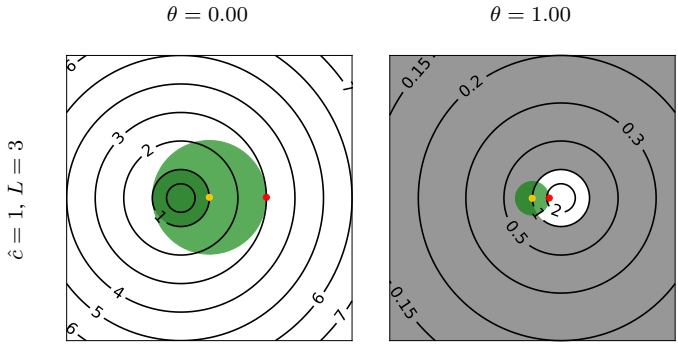
%\begin{figure}[t]
%    \centering
%    \stackinset{c}{0pt}{t}{-12pt}{\small $\theta=0.00$}{ \includegraphics[width=0.3\linewidth]{images/lip_stab_000.pdf}}
%	\stackinset{c}{0pt}{t}{-12pt}{\small $\theta=0.50$}{ \includegraphics[width=0.3\linewidth]{images/lip_stab_050.pdf}}
%	\stackinset{c}{0pt}{t}{-12pt}{\small $\theta=1.00$}{ \includegraphics[width=0.3\linewidth]{images/lip_stab_100.pdf}}
%	\\
%	\includegraphics[width=0.3\linewidth]{images/dissip_stab_000.pdf}
%	\includegraphics[width=0.3\linewidth]{images/dissip_stab_050.pdf}
%	\includegraphics[width=0.3\linewidth]{images/dissip_stab_100.pdf}
%    \caption{Contours of $|\mathcal{R}_{\theta}|$ and disks $\mathcal{B}\Big(\mathcal{R}^{-1}_{\theta}(\hat{c}),\mathcal{R}^{-1}_{\theta}(L)-\mathcal{R}^{-1}_{\theta}(\hat{c})\Big)$. Red points correspond to $\mathcal{R}^{-1}_{\theta}(\hat{c})$, $\mathcal{R}^{-1}_{\theta}(L)$ with (top) $\hat{c}=1$, $L=3$ and (bottom) $\hat{c}=0$, $L=1$. Note that $\sup_{\lambda\in\mathcal{B}}|\mathcal{R}_{\theta}(\lambda)|=L$.}
%    \label{fig:stab_regions_with_circle}
%\end{figure}

From the definition of spectral radius, we have
\begin{align*}
	\rho\left( \frac{\partial F(\tilde{\gamma},x)}{\partial x} \right)
	\leq \left\|\frac{\partial F(\tilde{\gamma},x)}{\partial x} \right\| \leq 1
	\quad\to\quad
	\lambda_i\in\mathcal{B}(0,1) \quad\forall i,
	%Re\big(\lambda_i\big) \in [-1,1] \quad\forall i,
\end{align*}
and thus all eigenvalues of $D_x F(\tilde{\gamma},x)$ are located in the unit circle $\mathcal{B}(0,1)\in\mathbb{C}$.
By denoting $\tilde{F}(\gamma,x):=F(\tilde{\gamma},x)$ and by appropriately scaling and shifting $\tilde{F}(\gamma,x)$, we redefine the vector field as
\begin{align*}
	F(\gamma,x) &:= c \cdot x + r \cdot \tilde{F}(\gamma,x)
	\quad\to\quad \lambda_i\in\mathcal{B}(c,r) \quad\forall i,
\end{align*}
so that all eigenvalues of $D_x F(\gamma,x)$ are now located in the disc $\mathcal{B}(c,r)$ with radius $r$ centered at~$c$.

%As Figure \ref{fig:stab_regions_with_circle} illustrates, 
It is convenient to define the disk $\mathcal{B}(c,r)$ in terms of the values of the stability function $\mathcal{R}_{\theta}$ as follows
\begin{align*}
	c &:= \mathcal{R}^{-1}_{\theta}(\hat{c}),
	\qquad
	r := \max\Big(0, \mathcal{R}^{-1}_{\theta}(L)-c \Big)
\end{align*}
with
\begin{alignat}{2}
	\label{eq:center}
	\hat{c}(\gamma_c) &:= \hat{c}_1 + S(\gamma_c) \cdot (\hat{c}_2-\hat{c}_1), \quad &&\hat{c}_i>0,
	\\
	\label{eq:Lipschitz}
	L(\gamma_L) &:= L_1 + S(\gamma_L) \cdot (L_2-L_1), \quad &&L_i>0
\end{alignat}
where $S(\cdot)$ is the sigmoid function, $\gamma_c$, $\gamma_L$ are the scalar valued parameters and $\mathcal{R}^{-1}_{\theta}$ is the inverse stability function
\begin{align*}%\label{eq:inv_stab_fun}
	\mathcal{R}_{\theta}(z) = \frac{1+(1-\theta)z}{1-\theta z}
	\quad\to\quad
	\mathcal{R}^{-1}_{\theta}(z) = \frac{1-z}{\theta(1-z)-1}.
\end{align*}
It can be shown that for $c>\mathcal{R}^{-1}_{\theta}(0)$ and $\lambda\in\mathcal{B}(c,r)$, $\mathcal{R}_{\theta}(\lambda)$ attains its maximum value at $\lambda$ with largest $Re(\lambda)$, see Figure~\ref{fig:stab_regions_with_circle} for illustration. 
This value defines the Lipschitz constant of a residual layer in \eqref{eq:layer_block} as follows
\begin{align*}%\label{eq:resid_lipchitz}
%	Lip(\Phi_{T,\theta}) 
	\sup_{\lambda\in\mathcal{B}(c,r)} |\mathcal{R}_{\theta}(\lambda)|
	= \max\big(\hat{c}(\gamma_c),L(\gamma_L)\big).
\end{align*}
The above expression allows to explicitly bound the Lipschitz constant within the given range by means of $\hat{c}_i,L_i$ in \eqref{eq:center}-\eqref{eq:Lipschitz} and also to include it into the optimization problem through learnable parameters $\gamma_c$, $\gamma_L$.
%Note that by initializing $S(\gamma_c)=S(\gamma_L)\approx 0$, we can interpret $\hat{c}_1$, $L_1$ and $\hat{c}_2$, $L_2$ as the initial and possible final values for $\hat{c}(\gamma_c)$, $L(\gamma_L)$ respectively.

%\begin{definition}\label{def:bounded_layer}
%We say that the residual layer in \eqref{eq:layer_block} with the vector field $F^L(\gamma,x)$ %that has parameterization 
%%$$F^L(\gamma,x): \quad c:=0, \; r := \max\Big(0, \mathcal{R}^{-1}_{\theta}\big(2 + S(\gamma_L) \cdot (L-2)\big) \Big)$$
%$$F^L(\gamma,x): \quad c:=0, \; r := \max\Big(0, \mathcal{R}^{-1}_{\theta}\big(1 + S(\gamma_L) \cdot (L-1)\big) \Big)$$
%is an \textbf{$L$-bounded residual layer}. % and it is denoted by $\Phi^L_T$.
%%We define the \textbf{bounded residual layer} as follows
%%\begin{align*} %\label{eq:dissip_block}
%%	\Phi^L_T:
%%	\qquad
%%    \begin{cases}
%%	    y_0 = x,
%%	    \\
%%	    y_t = y_{t-1} + F^L(\gamma_0,y_{t-1}),
%%    \end{cases}
%%    \qquad t=1,\hdots,T,
%%\end{align*}
%%where the vector field $F^L$ has parameterization $$c:=0, \quad r := \max\Big(0, \mathcal{R}^{-1}_{\theta}\big(2 + S(\gamma_L) \cdot (L-2)\big) \Big).$$ %L(\gamma_L) := 2 + S(\gamma_L) \cdot (L-2).$$
%We denote such layers by $\Phi^L$.
%It is clear that the Lipschitz constant of $\Phi^L$ is in $[1,L]$.
%\end{definition}

\begin{theorem}\label{th:dissip_params}
The residual layer is dissipative %in \eqref{eq:layer_block} 
if its vector field is parameterized as %$\hat{F}(\gamma,x)$
$$F(\gamma,x) := c \cdot x + r \cdot \tilde{F}(\gamma,x)$$
% is parameterized as
with
%\begin{align*}
%	\hat{F}(\gamma,x) &:= c \cdot x + r \cdot \tilde{F}(\gamma,x)
%\end{align*}
%is dissipative if
%One can define the dissipative residual layer by setting
\begin{align*}%\label{eq:dissip_params}
%	\hat{F}(\gamma,x):
%	\quad
	&c := \mathcal{R}^{-1}_{\theta}\big(1-S(\gamma_c)\big),
	\\
	&r := \max\Big(0, \mathcal{R}^{-1}_{\theta}\big(1-S(\gamma_L)\big)-c \Big).
\end{align*}
%i.e., both $\hat{c}\in(0,1)$ and $L\in(0,1)$.
\end{theorem}
\begin{proof}
Follows directly from Definition~\ref{def:dissip_layer} since both $\hat{c},L\in(0,1)$.
\end{proof}

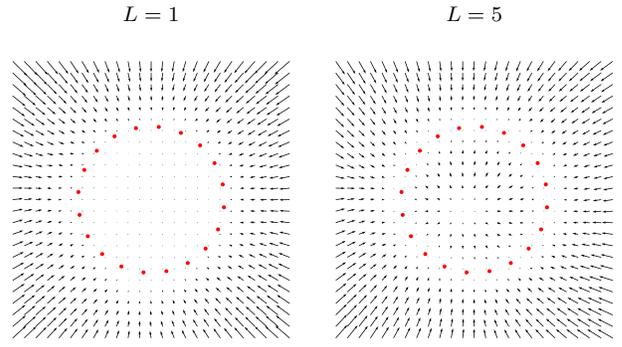
\begin{figure}[t]
 	\centering
	\pgfplotsset{every tick label/.append style={font=\scriptsize}}
	\pgfplotsset{every axis label/.append style={font=\footnotesize}}
	\pgfplotsset{every axis title/.append style={font=\footnotesize}} 	
 	
 	\foreach \L in {1, 5}
 	{
			\begin{tikzpicture}[trim axis left,trim axis right]
 				\begin{axis} [
 					width=0.36\textwidth,
 					ymode = normal,
 					xticklabels={,,},
 					yticklabels={,,},
 					enlargelimits=false,
 					axis equal image,
 					axis on top,
 					axis line style={draw=none},
 					title={$L=\L$},
 					%label=\ifthenelse{\equal{\th}{000}}{$\hat{c}=\c$, $L=\L$}{},
 					tick style={draw=none},
 					very thick ]
 					\addplot graphics [points={(-5,-5) (5,5)}, includegraphics={trim=0 0 0 0,clip}] {{{images/circle_vector_field_lip\L}.pdf}};
 				\end{axis}
 			\end{tikzpicture}
	}
	\caption{Vector fields of a residual layer trained with different bounds on the Lipschitz constant.}
    \label{fig:circle_lip} 
\end{figure}

\textbf{Remark.}
Dissipative layers have the obvious benefit of being unconditionally stable uniformly for all inputs.
However, Figure~\ref{fig:circle_lip} shows the situation when the dissipativity condition becomes overly restrictive.
The vector fields in this example were trained to vanish on the boundary of the circle while being locally attractive to it.
According to the Gauss's theorem, the total of the sources in the closed volume is equal to the flux of the vector field across the boundary.
Since the vector field vanishes on the boundary, it must have sources inside the volume to satisfy the condition of local attractiveness.
To get nontrivial solutions, we are forced to allow for the Lipschitz constant of the layer to be greater than one to include the region of positive divergence.
In all examples below, we will use $L\in[1,5]$.

%\begin{definition}\label{def:stabilized_layer}
%We define the \textbf{stabilized residual layer} as follows
%\begin{align*} %\label{eq:dissip_block}
%	\hat{\Phi}^L(x):
%	\qquad
%    \begin{cases}
%	    \hat{y} = x + F^L(\gamma_0,x),
%	    \\
%	    y = \hat{y} + \hat{F}(\gamma_t,\hat{y}).
%    \end{cases}
%\end{align*}
%We will also use more general case with implicit dissipative part defined as
%\begin{align*} %\label{eq:dissip_block}
%	%RL^L_{T,\theta}:
%	\hat{\Phi}^L_{\theta}(x):
%	\qquad
%    \begin{cases}
%	    %y_0 = x,
%	    %\\
%	    \hat{y} = x + F^L(\gamma_0,x),
%	    \\
%	    y = \hat{y} + \hat{F}(\gamma_t,(1-\theta)\hat{y}+\theta y).
%    \end{cases}
%    %\qquad t=1,\hdots,T,
%\end{align*}
%%where $\hat{F}$ is the vector field of a dissipative bottleneck.
%Similarly to $\Phi^L$, the Lipschitz constant of $\hat{\Phi}^L$/$\hat{\Phi}^L_{\theta}$ is in $[2,L]$.
%\end{definition}

\subsection{Dissipative manifolds}
\label{sec:reg}

\begin{figure*}[t]
 	\centering
	\pgfplotsset{every tick label/.append style={font=\scriptsize}}
	\pgfplotsset{every axis label/.append style={font=\footnotesize}}
	\pgfplotsset{every axis title/.append style={font=\footnotesize}} 	
 	
 	\foreach \s in {0,5,10,20}
 	{
			\begin{tikzpicture}[trim axis left,trim axis right]
 				\begin{axis} [
 					width=0.33\textwidth,
 					ymode = normal,
 					xticklabels={,,},
 					yticklabels={,,},
 					enlargelimits=false,
 					axis equal image,
 					axis on top,
 					axis line style={draw=none},
 					title={$t=\s$},
 					%label=\ifthenelse{\equal{\th}{000}}{$\hat{c}=\c$, $L=\L$}{},
 					tick style={draw=none},
 					very thick ]
 					\addplot graphics [points={(-5,-5) (5,5)}, includegraphics={trim=0 0 0 0,clip}] {{{images/scurve_points_\s_test}.pdf}};
 				\end{axis}
 			\end{tikzpicture}
	}
	\caption{One-dimensional dissipative manifold in two-dimensional space.}
    \label{fig:scurve_evol} 
\end{figure*}
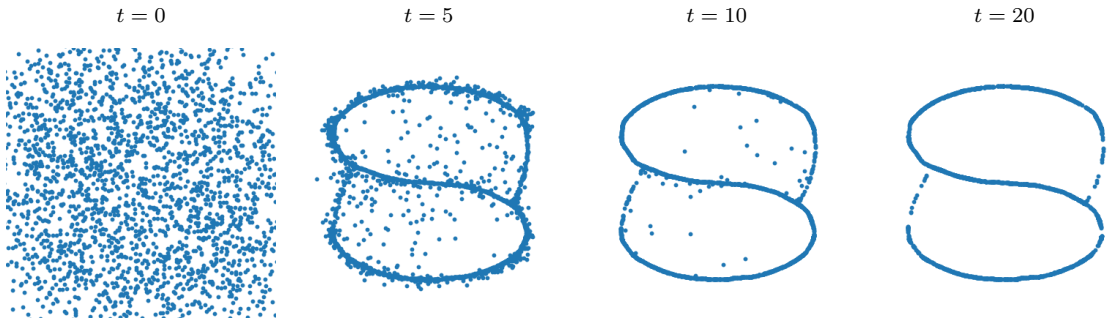

\begin{definition}\label{def:dissip_manifold}
We define the \textbf{dissipative manifold} $\mathcal{M}_d$ as a set of points that has vanishing vector field and is locally attractive.
In other words, it is the level set $$\mathcal{M}_d=\{x:\|F(\gamma,x)\|^2=0, |\mathcal{R}_{\theta}(\lambda_i)|<1\}$$ for all eigenvalues $\lambda_i$ of the Jacobian $D_xF(\gamma,x)$, $x\in\mathcal{M}_d$.
\end{definition}

According to the above definition, any point on the manifold must be stationary, and any point close enough to the manifold should be transported back to the manifold by the flow of the corresponding residual layer.
%The choice of parameterization in Theorem~\ref{th:dissip_params} ensures dissipativity of a residual layer.
%However, it doesn't impose any constraints on the location of the spectrum within the disc $\mathcal{B}(c,r)$.

The first goal can be achieved by regularizing the magnitude of the vector field 
\begin{align}\label{eq:F_reg}
	R_F(\gamma) := \|F(\gamma,x)\|^2.
\end{align}
To achieve the second goal, for every data point on the manifold, we are interested in concentrating as many eigenvalues of the corresponding Jacobian as possible in the vicinity of $\mathcal{R}^{-1}_{\theta}(0)$ so that the corresponsing degrees of freedom effectively vanish.
For this purpose, use the following bound 
%We can further restrict the spectrum of a vector field $F(\gamma,y)$ by considering the divergence regularization of the form
%In order to reduce the 
%In addition to the spectral localization in \eqref{eq:our_F}, we consider the divergence regularization of the form
%Instead, we consider ``discrete-time" regularization of the form
\begin{align*}
	&\sum_i \left| \lambda_i \Big( \mathcal{R}_{\theta}(D_xF) \Big) \right|^2
	\leq \sum_i \sigma_i^2 \Big( \mathcal{R}_{\theta}(D_xF) \Big)
	\\
	&= \left\| \mathcal{R}_{\theta}(D_xF) \right\|_F^2
	= \Tr \Big( \mathcal{R}_{\theta}(D_xF)^T \mathcal{R}_{\theta}(D_xF) \Big)
\end{align*}
and consider the trace regularization of the form
\begin{align}\label{eq:eig_reg}
	R_{\lambda}(\gamma) :=
	\Tr \Big( \mathcal{R}_{\theta}(D_xF)^T \mathcal{R}_{\theta}(D_xF) \Big).
\end{align}
The Frobenius norm of the Jacobian can be efficiently estimated using the stochastic Hutchinson estimator, see \cite{reshniak2020robust} for an example.
%\begin{align}\label{eq:our_reg}
%	R(\gamma) :=
%	\mathcal{R}_{\theta}
%	\left(
%	\frac{1}{dT}
%	\sum_{t=1}^T \nabla \cdot F(\gamma,(1-\theta)y_{t-1}+\theta y_t)
%	\right),
%\end{align}
%where $d$ is the dimension of the vector field $F:\mathbb{R}^p\times\mathbb{R}^d \to \mathbb{R}^d$. %, and $\nabla \cdot F(\gamma,y)$ is its divergence.

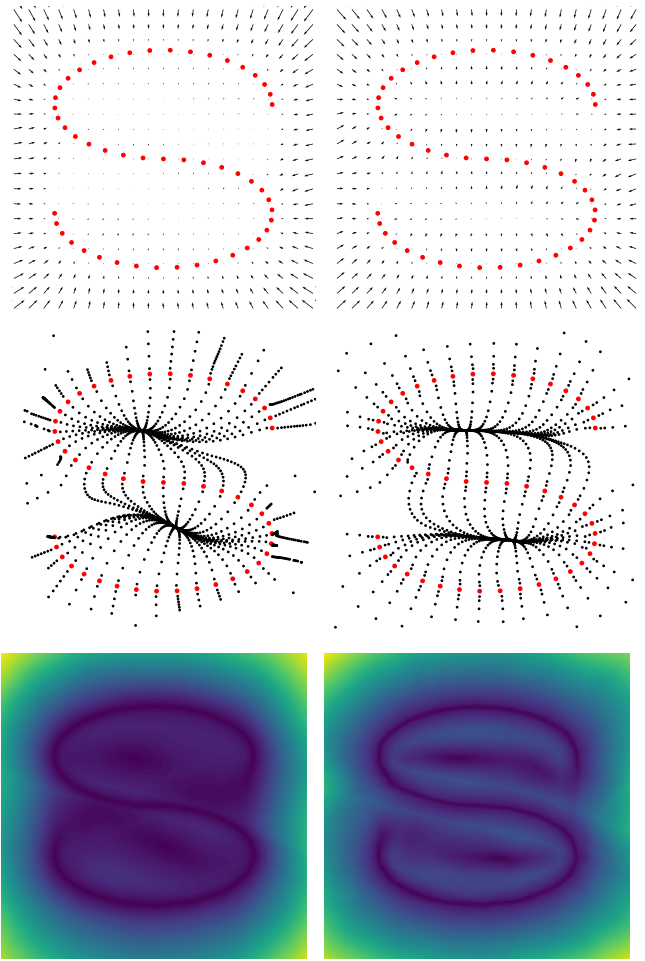
\begin{figure}[t]
 	\centering
	\pgfplotsset{every tick label/.append style={font=\scriptsize}}
	\pgfplotsset{every axis label/.append style={font=\footnotesize}}
	\pgfplotsset{every axis title/.append style={font=\footnotesize}} 	
 	
 	\foreach \s in {1,3}
 	{
			\begin{tikzpicture}[trim axis left,trim axis right]
 				\begin{axis} [
 					width=0.36\textwidth,
 					ymode = normal,
 					xticklabels={,,},
 					yticklabels={,,},
 					enlargelimits=false,
 					axis equal image,
 					axis on top,
 					axis line style={draw=none},
 					%title={$L=\L$},
 					%label=\ifthenelse{\equal{\th}{000}}{$\hat{c}=\c$, $L=\L$}{},
 					tick style={draw=none},
 					very thick ]
 					\addplot graphics [points={(-5,-5) (5,5)}, includegraphics={trim=0 0 0 0,clip}] {{{images/scurve_vector_field_\s}.pdf}};
 				\end{axis}
 			\end{tikzpicture}
	}
	\\
 	\foreach \s in {1,3}
 	{
			\begin{tikzpicture}[trim axis left,trim axis right]
 				\begin{axis} [
 					width=0.36\textwidth,
 					ymode = normal,
 					xticklabels={,,},
 					yticklabels={,,},
 					enlargelimits=false,
 					axis equal image,
 					axis on top,
 					axis line style={draw=none},
 					%title={$L=\L$},
 					%label=\ifthenelse{\equal{\th}{000}}{$\hat{c}=\c$, $L=\L$}{},
 					tick style={draw=none},
 					very thick ]
 					\addplot graphics [points={(-5,-5) (5,5)}, includegraphics={trim=0 0 0 0,clip}] {{{images/scurve_traj_\s}.pdf}};
 				\end{axis}
 			\end{tikzpicture}
	}
	\\
 	\foreach \s in {1,3}
 	{
			\begin{tikzpicture}[trim axis left,trim axis right]
 				\begin{axis} [
 					width=0.36\textwidth,
 					ymode = normal,
 					xticklabels={,,},
 					yticklabels={,,},
 					enlargelimits=false,
 					axis equal image,
 					axis on top,
 					axis line style={draw=none},
 					%title={$L=\L$},
 					%label=\ifthenelse{\equal{\th}{000}}{$\hat{c}=\c$, $L=\L$}{},
 					tick style={draw=none},
 					very thick ]
 					\addplot graphics [points={(-5,-5) (5,5)}, includegraphics={trim=0 0 0 0,clip}] {{{images/scurve_contours_\s}.pdf}};
 				\end{axis}
 			\end{tikzpicture}
	}
	\caption{Vector fields, adjoint trajectories, and the level sets of the 2d curve for 1 (left) and 3 (right) steps of the adjoint solver in regularizer \eqref{eq:adjoint_reg}.}
    \label{fig:scurve} 
\end{figure}

Regularizers in \eqref{eq:F_reg}-\eqref{eq:eig_reg} are local to the data manifold since they only require original data points.
Better results can be obtained by considering nonlocal perturbations in the directions orthogonal to the manifold.
These orthogonal directions can be obtained trivially by computing the gradient of the level set at each point as $n=\nabla\|F(\gamma,x+\epsilon)\|^2$. The small random perturbation $\epsilon$ is chosen to ensure the uniqueness of the gradient.
The corresponding regularization for the explicit residual layer is then given by
\begin{align}\label{eq:ortho_reg}
	R_n(\gamma) = \| F(\gamma,x+\alpha n) + \alpha n \|^2
\end{align}
for some small $\alpha>0$.
More perturbation points can be obtained by stepping backwards in time along the trajectory ending at $x+\alpha n$.
This gives
\begin{align}\label{eq:adjoint_reg}
	R_{adj}(\gamma) = \sum_{j} \| F(\gamma,x_j) + x_j - x \|^2,
\end{align}
where $x_j$ is the trajectory generated by the adjoint solver, e.g., $$y = x + F(x) \quad\to\quad x = y - F(x).$$
Figure~\ref{fig:scurve} illustrates the impact of the number of steps in \eqref{eq:adjoint_reg} on the learned dynamics.
%One can see that the result of using more steps is more favorable in terms of 
Adding more steps in \eqref{eq:adjoint_reg} favors more aggressive exploration of the space around the manifold.
One can see from the Figure that this can result in a better adaptation to the geometry of the data.
Figure~\ref{fig:scurve_evol} also shows the learned evolution of the random point cloud that is attracted to the original data manifold.

\section{Example}

As a final example demonstrating the properties of the proposed manifold representation approach, consider the digits dataset from UCI ML Repository \cite{alpaydin1998optical}.
This dataset contains 5620 8x8 pixel 16 bit images of handwritten digits each treated as a vector of size 64.

We parameterized the vector field of the residual layer by a ReLU network of depth $2$ and width $1000$.
We trained the explicit ($\theta=0$) residual layer by minimizing the regularizers in \eqref{eq:F_reg} and \eqref{eq:eig_reg} using Adam optimizer with step size $10^{-3}$ for 10000 epochs.

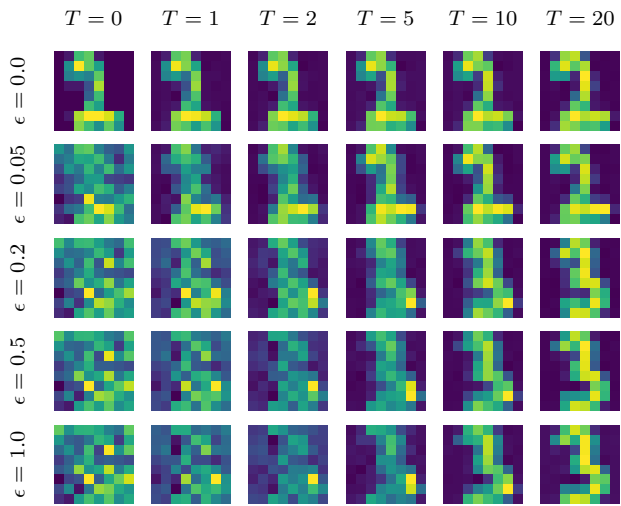
\begin{figure}[t]
 	\centering
	\pgfplotsset{every tick label/.append style={font=\scriptsize}}
	\pgfplotsset{every axis label/.append style={font=\footnotesize}}
	\pgfplotsset{every axis title/.append style={font=\footnotesize}} 	
 	
 	\foreach \lip in {5}
 	{
% 		\foreach \eps/\epss in {000/0.00,001/0.01,005/0.05,010/0.10,020/0.20,050/0.50,100/1.00}
 		\foreach \eps/\epss in {000/0.0,005/0.05,020/0.2,050/0.5,100/1.0}
 		{
		\foreach \T in {0,1,2,5,10,20}
		{
			\begin{tikzpicture}[trim axis left,trim axis right]
 				\begin{axis} [
 					width=0.145\textwidth,
 					height=0.145\textwidth,
 					ymode = normal,
 					xticklabels={,,},
 					yticklabels={,,},
 					enlargelimits=false,
 					%axis equal image,
 					axis on top,
 					axis line style={draw=none},
 					title=\ifthenelse{\equal{\eps}{000}}{$T=\T$}{},
 					ylabel=\ifthenelse{\equal{\T}{0}}{$\epsilon=\epss$}{},
 					tick style={draw=none},
 					very thick ]
 					\addplot graphics [points={(0,0) (8,8)}, includegraphics={trim=0 0 0 0,clip}] {{{images/digit_501_T\T_eps\eps_clip0_265_1000_2_000_5}.pdf}};
 				\end{axis}
 			\end{tikzpicture}
 		}
 		\\[-1.5em]
 		}
	}
%	\vspace{2em}
	\caption{Denoising of a digit corrupted with additive Gaussian noise.} % of various intensity}
    \label{fig:digits_noclip_gaus} 
\end{figure}

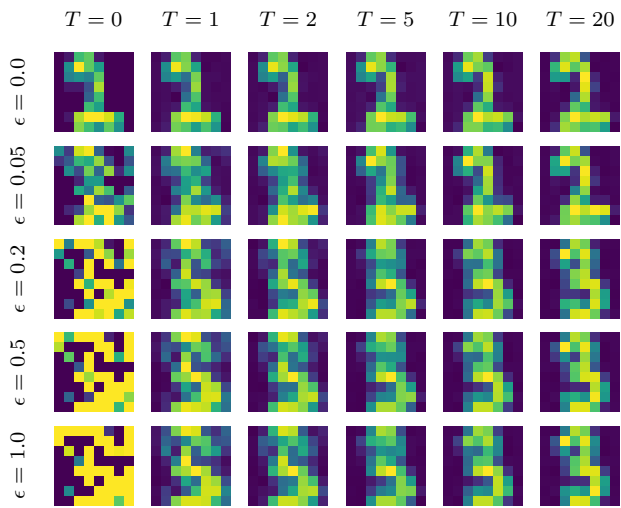
\begin{figure}[t]
 	\centering
	\pgfplotsset{every tick label/.append style={font=\scriptsize}}
	\pgfplotsset{every axis label/.append style={font=\footnotesize}}
	\pgfplotsset{every axis title/.append style={font=\footnotesize}} 	
 	
 	\foreach \lip in {5}
 	{
% 		\foreach \eps/\epss in {000/0.00,001/0.01,005/0.05,010/0.10,020/0.20,050/0.50,100/1.00}
 		\foreach \eps/\epss in {000/0.0,005/0.05,020/0.2,050/0.5,100/1.0}
 		{
		\foreach \T in {0,1,2,5,10,20}
		{
			\begin{tikzpicture}[trim axis left,trim axis right]
 				\begin{axis} [
 					width=0.145\textwidth,
 					height=0.145\textwidth,
 					ymode = normal,
 					xticklabels={,,},
 					yticklabels={,,},
 					enlargelimits=false,
 					%axis equal image,
 					axis on top,
 					axis line style={draw=none},
 					title=\ifthenelse{\equal{\eps}{000}}{$T=\T$}{},
 					ylabel=\ifthenelse{\equal{\T}{0}}{$\epsilon=\epss$}{},
 					tick style={draw=none},
 					very thick ]
 					\addplot graphics [points={(0,0) (8,8)}, includegraphics={trim=0 0 0 0,clip}] {{{images/digit_501_T\T_eps\eps_clip1_265_1000_2_000_5}.pdf}};
 				\end{axis}
 			\end{tikzpicture}
 		}
 		\\[-1.5em]
 		}
	}
%	\vspace{2em}
	\caption{Denoising of a digit corrupted with truncated Gaussian noise.} % of various intensity}
    \label{fig:digits_clip_gaus} 
\end{figure}

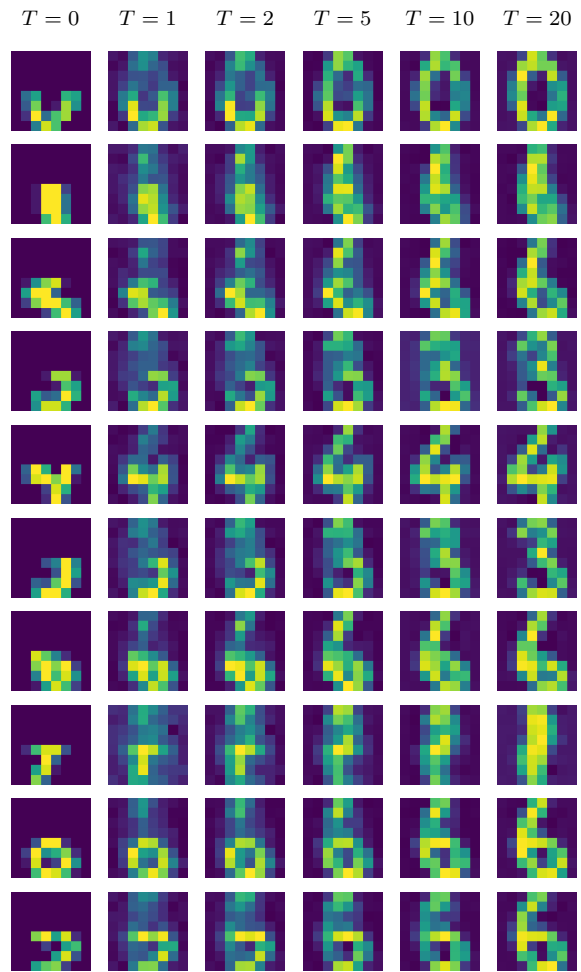
\begin{figure}[t]
 	\centering
	\pgfplotsset{every tick label/.append style={font=\scriptsize}}
	\pgfplotsset{every axis label/.append style={font=\footnotesize}}
	\pgfplotsset{every axis title/.append style={font=\footnotesize}} 	
 	
 	\foreach \lip in {5}
 	{
% 		\foreach \eps/\epss in {000/0.00,001/0.01,005/0.05,010/0.10,020/0.20,050/0.50,100/1.00}
 		\foreach \digit in {0,...,9}
 		{
		\foreach \T in {0,1,2,5,10,20}
		{
			\begin{tikzpicture}[trim axis left,trim axis right]
 				\begin{axis} [
 					width=0.145\textwidth,
 					height=0.145\textwidth,
 					ymode = normal,
 					xticklabels={,,},
 					yticklabels={,,},
 					enlargelimits=false,
 					%axis equal image,
 					axis on top,
 					axis line style={draw=none},
 					title=\ifthenelse{\equal{\digit}{0}}{$T=\T$}{},
 					%ylabel=\ifthenelse{\equal{\T}{0}}{\digit}{},
 					tick style={draw=none},
 					very thick ]
 					\addplot graphics [points={(0,0) (8,8)}, includegraphics={trim=0 0 0 0,clip}] {{{images/digit_\digit_T\T_half_265_1000_3_000_5_relu}.pdf}};
 				\end{axis}
 			\end{tikzpicture}
 		}
 		\\[-1.5em]
 		}
	}
%	\vspace{2em}
	\caption{Generating digits from missing data}
    \label{fig:digits_generative} 
\end{figure}

Figures~\ref{fig:digits_noclip_gaus} and \ref{fig:digits_clip_gaus} show the outcome of the trained layer for the inputs corrupted with additive and truncated Gaussian noise.
One can see that for the noise with variance $\epsilon=0.05$, the layer was able to reproduce the original image very accurately even though the corrupted image is difficult to recognize visually.
For the larger levels of noise, the layer did not reproduce the original image but still produced a very clear output of a different digit.
This indicates that the learned vector field performs projection on the original data manifold as desired.
Figure~\ref{fig:digits_generative} supports this finding by illustrating the completion of partially missing data.
It shows that the proposed data manifold representation approach can be also used for generative tasks.

%\begin{figure*}[t]
% 	\centering
%	\pgfplotsset{every tick label/.append style={font=\scriptsize}}
%	\pgfplotsset{every axis label/.append style={font=\footnotesize}}
%	\pgfplotsset{every axis title/.append style={font=\footnotesize}} 	
% 	
% 	\foreach \lip in {5}
% 	{
% 		\foreach \depth in {0,1,2}
% 		{
%		\foreach \T in {0,1,2,5,10,20}
%		{
%			\begin{tikzpicture}[trim axis left,trim axis right]
% 				\begin{axis} [
% 					width=0.2\textwidth,
% 					height=0.2\textwidth,
% 					ymode = normal,
% 					xticklabels={,,},
% 					yticklabels={,,},
% 					enlargelimits=false,
% 					%axis equal image,
% 					axis on top,
% 					axis line style={draw=none},
% 					title=\ifthenelse{\equal{\depth}{0}}{$T=\T$}{},
% 					ylabel=\ifthenelse{\equal{\T}{0}}{$depth=\depth$}{},
% 					tick style={draw=none},
% 					very thick ]
% 					\addplot graphics [points={(0,0) (8,8)}, includegraphics={trim=0 0 0 0,clip}] {{{images/digit_501_T\T_eps005_clip0_265_1000_\depth_000_5}.png}};
% 				\end{axis}
% 			\end{tikzpicture}
% 		}
% 		\\%[-1.5em]
% 		}
%	}
%	\vspace{2em}
%	\caption{width 1000, $\epsilon=0.05$, noclip}%; (fourth row) evolution of the training loss with $95\%$ confidence intervals over $10$ independent runs.}
%%%	\caption{Learned vector fields for the problem in Example~\ref{ex:blobs} for bounded and stabilized residual layers with and without trace regularization.}
%%    \label{fig:blobs_results} 
%\end{figure*}

%\subsection{MNIST}
%\label{ex:mnist}

\section{Conclusions and future work}
A method to construct an implicit parameterization of data manifolds was proposed.
%It does not require data augmentation 
It is end-to-end trainable and can be easily added to any existing network.
Performance of the method was demonstrated on denoising and generative tasks.
The obtained results are encouraging but more efforts are required for the detailed comparison with existing alternatives.
Potential applications include those that can benefit from stabilizing their latent dynamics. % and are also yet to be explored.
The problem of stabilizing optimal policies in the tasks of reinforcement learning is of particular interest and will be studied in future works.
%
% finding a stable optimal policy 
%
%

%\section*{References}
\bibliographystyle{plain}
\bibliography{biblio}

\end{document}